\documentclass[journal]{IEEEtran}

\usepackage{cite}
\usepackage{amsmath,amssymb,amsfonts}
\usepackage{bm}
\usepackage{graphicx}
\usepackage{xcolor}
\usepackage{booktabs}
\usepackage{array}
\usepackage{tabularx}
\usepackage{tikz}
\usetikzlibrary{arrows.meta,positioning,fit}
\usepackage{placeins}
\usepackage{stfloats}
\usepackage[hidelinks]{hyperref}
\hypersetup{
  pdftitle={EndoLIFT: Language-Disambiguated Latent-Conditioned Rectified Flow for Bidirectional Endoscopic Control},
  pdfauthor={Chi Kit Ng, Yidong Zhang, Lui Siu Hing, Jinsong Lin, Tianchun Wu, Ho Yin Chim, Zhiqing Tang, Tao Yang, Huxin Gao, Trevor Yeung, Raymond Shing-Yan Tang, and Hongliang Ren}
}

\graphicspath{{figures/}}
\newcommand{\methodname}{EndoLIFT}
\newcommand{\latentname}{variational trajectory latent}
\newcommand{\VTL}{VTL}
\newcommand{\lcrf}{latent-conditioned rectified flow}

\newcommand{\dinobaseline}{DINOv2 + MLP}
\newcommand{\qwenmlp}{Qwen3-VL + MLP}
\newcommand{\qwengroot}{Qwen3-VL + GR00T}
\newcommand{\rfwithoutlatent}{EndoLIFT w/o VTL}
\newcommand{\ourslabel}{EndoLIFT (ours)}
\newcommand{\secref}[1]{Section~\ref{#1}}
\newcommand{\figref}[1]{Fig.~\ref{#1}}
\newcommand{\tabref}[1]{Table~\ref{#1}}
\renewcommand{\arraystretch}{1.08}

\title{EndoLIFT: Language-Disambiguated Latent-Conditioned Rectified Flow for Bidirectional Endoscopic Control}

\author{Chi Kit Ng\textsuperscript{*}, Yidong Zhang\textsuperscript{*},
Lui Siu Hing\textsuperscript{*}, Jinsong Lin\textsuperscript{*}, Tianchun Wu,
Ho Yin Chim, Zhiqing Tang, Tao Yang, Huxin Gao, Trevor Yeung,
Raymond Shing-Yan Tang, and Hongliang Ren%
\thanks{\textsuperscript{*}Chi Kit Ng, Yidong Zhang, Lui Siu Hing, and
Jinsong Lin contributed equally to this work.}%
\thanks{Chi Kit Ng, Yidong Zhang, Lui Siu Hing, Jinsong Lin, Tianchun Wu,
Ho Yin Chim, Zhiqing Tang, Huxin Gao, Trevor Yeung, Raymond Shing-Yan Tang,
and Hongliang Ren are with The Chinese University of Hong Kong, Hong Kong SAR,
China.}%
\thanks{Tao Yang is with The Sixth Affiliated Hospital, Sun Yat-sen
University, Guangzhou, China.}}

\begin{document}
\maketitle
\raggedbottom

\begin{abstract}
Routine gastrointestinal endoscopy is intrinsically bidirectional: the
instrument is advanced to reach target anatomy and later withdrawn or
retroflexed for inspection, while an external cue may require earlier reversal.
When the requested phase changes before the visual scene does, nearly identical
observations can require opposite axial actions. We identify and formalize this
ambiguity in bidirectional endoscopic control as \emph{intent aliasing}. We
propose EndoLIFT (\emph{Endoscopic Language-Instruction Flow with Trajectory
Latents}), a vision-language-action policy that combines explicit
language-based intent conditioning with a latent-conditioned rectified-flow
action expert. The policy receives RGB, a language instruction, and the
previous-action state; a 32-D variational trajectory latent stochastically
conditions continuous action-chunk generation. Controlled same-observation
instruction swaps establish that language selects the axial mode, independently
of whether the trajectory latent is present. Relative to the matched model
without latent conditioning, \methodname{} improves navigation-direction
accuracy by 11.1 percentage points and reduces wrong-direction advance by 83\%.
An architecture-controlled 1-bit mode-flag reference exhibits weaker canonical-anchor switching,
while \methodname{} retains 82.8\% intent-following accuracy across 44 held-out linguistic variants. In
closed-loop evaluation, \methodname{} improves overall success by 30 percentage
points over \rfwithoutlatent{} on both the seen colon phantom and the unseen
lung and stomach phantoms, and completes 10/10 ex-vivo porcine-trachea trials.
These results separate language-based intent selection from the trajectory
latent's contribution to directional correctness and robust retraction.
\end{abstract}


\section{Introduction}
\label{sec:intro}

Gastrointestinal examinations are organized around complementary motion
phases rather than a single forward trajectory. In colonoscopy, the endoscope
is advanced to the cecum and detailed mucosal inspection is performed during
withdrawal; withdrawal quality is associated with lesion
detection~\cite{barclay2006withdrawal}. Systematic upper-gastrointestinal
examination similarly requires ordered inspection of anatomical landmarks in
antegrade and retroflexed views~\cite{emura2020photodocumentation}. This
insertion--inspection workflow has also motivated robotic endoscopy systems
that explicitly reproduce insertion and withdrawal
phases~\cite{xu2022bidirectional}. A useful autonomous endoscope must therefore
advance through deformable anatomy, maintain a usable view, and reverse in a
controlled manner when the procedural phase changes. Reverse motion may also
be requested earlier by an urgent physiological event or a verified operator
command.

\begin{figure*}[!t]
\centering
\includegraphics[width=\textwidth,trim=0 190bp 0 190bp,clip]{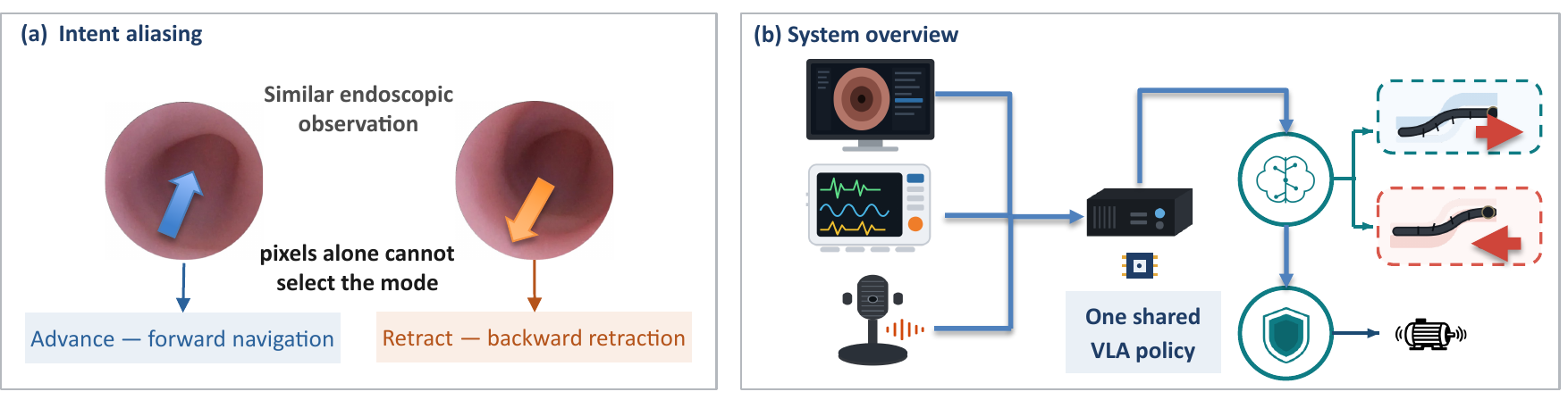}
\caption{Motivation and system concept. (a) \emph{Intent aliasing}: visually
similar endoscopic observations can require opposite axial actions, so pixels
alone cannot select the active procedural mode. (b) An external monitor--speech
supervisor selects a canonical forward or retraction instruction for one shared
VLA policy, which maps the image, instruction, and previous action to
three-axis endoscope motion.}
\label{fig:overview}
\end{figure*}

The required phase is not always inferable from pixels. At the instant an
operator or monitor requests retraction, the endoscopic image may have barely
changed and continued forward motion may remain geometrically feasible, yet it
is now procedurally wrong. \figref{fig:overview}(a) illustrates how visually
neighboring observations can therefore require opposite axial actions; we call this
\emph{intent aliasing}. Without an explicit mode variable, the
observation-to-action map is multi-valued: squared-error imitation can average
opposing actions or follow the dominant mode~\cite{florence2022ibc}, while a
generative policy can represent multiple action modes but still lacks evidence
about which mode should be executed. The missing variable is not another image
feature; it is the externally specified procedural intent.

In current clinical practice, flexible endoscopy remains predominantly
manually controlled, and end-to-end language-conditioned control is not part
of routine operation. During a procedure, the clinician's hands are already
occupied with manipulating the endoscope and accessories, whereas speech
remains readily available as a low-burden input channel. Language therefore
provides a practical and auditable interface for specifying high-level
procedural intent without requiring an additional hand-operated control
device. Language-conditioned robot policies connect task semantics with
perception and action~\cite{jang2022bcz,shridhar2022cliport,mees2022calvin,
jiang2023vima}, while large-scale embodied models and datasets broaden transfer
across tasks and robot embodiments~\cite{brohan2023rt1,oneill2024openx,
zitkovich2023rt2,ghosh2024octo,kim2025openvla}. Continuous VLA policies
additionally generate temporally coherent action
chunks~\cite{black2025pi0}. Endoscopy-specific VLAs have demonstrated
prompt-conditioned tracking, stage-aware navigation, and grounded
world--action modeling~\cite{ng2025endovla,lin2026bilivla,lin2026endowam};
Open-H-Embodiment further targets large-scale medical-robotics foundation-model
training~\cite{openh2026embodiment}. We study the complementary case in which
an external event changes the required axial mode before the visual scene
changes. In this setting, generative action modeling alone is insufficient:
the policy must also identify which externally specified procedural intent
should govern the current observation.

We propose \methodname{} (\emph{Endoscopic Language-Instruction Flow
with Trajectory Latents}; \figref{fig:overview}(b)), a vision-language-action
policy for language-disambiguated bidirectional endoscopic control. The neural
policy takes RGB, the canonical language instruction, and the previous executed
action. A PaliGemma~2 vision-language backbone provides multimodal context to a
Transformer action expert, while a 32-D \VTL{} stochastically conditions
\lcrf{} generation of continuous action chunks. An external trigger supervisor maps
physiological abnormalities and verified speech events to an instruction and
remains outside the learned network. The experiments focus on switching between
normal forward navigation and externally triggered urgent retraction.

The deployed supervisor uses two canonical instruction anchors, but the final
checkpoint is not trained on only those two strings: its base imitation table
contains four surface forms and a subsequent instruction-conditioning stage
adds paraphrases and perturbations. We make this distinction explicit and test
whether a literal 1-bit interface can replace text conditioning when the
upstream mode is already known.

Under the six-level medical-robotics taxonomy of Yang
et al.~\cite{yang2017medical}, the present prototype is best characterized as
\emph{Level~2 task autonomy} for bounded endoscopic motion: the operator
initiates or verifies the selected forward/retract task, the robot executes
continuous motion under monitoring, and the operator retains discrete
intervention authority. The system does not autonomously define clinical goals,
generate a complete procedural strategy, or claim whole-examination autonomy.

We first test whether language changes the axial mode while the image and
previous action remain fixed, then compare latent conditioning with a matched
rectified-flow ablation. Closed-loop trials use the colon phantom as the seen
domain and the lung and stomach phantoms as unseen luminal transfer tests.

This work makes three contributions:
\begin{enumerate}
  \item We identify and formalize \emph{intent aliasing} in bidirectional
  endoscopic control: nearly identical observations require opposite axial
  actions when the externally specified procedural intent changes.
  \item We introduce \methodname{}, a PaliGemma~2-based three-axis VLA whose
  32-D \emph{\latentname} (\VTL) stochastically conditions a rectified-flow
  Transformer action expert to generate $32{\times}3$ continuous action chunks
  from visual, language, and previous-action context.
  \item We isolate the intent interface and trajectory latent through
  same-observation instruction flips, an architecture-controlled ModeFlag reference,
  a matched no-VTL ablation, 44 held-out linguistic variants, cross-phantom transfer, and quantitative
  ex-vivo validation in a porcine trachea.
\end{enumerate}

\begin{figure*}[!t]
\centering
\includegraphics[width=0.8\textwidth]{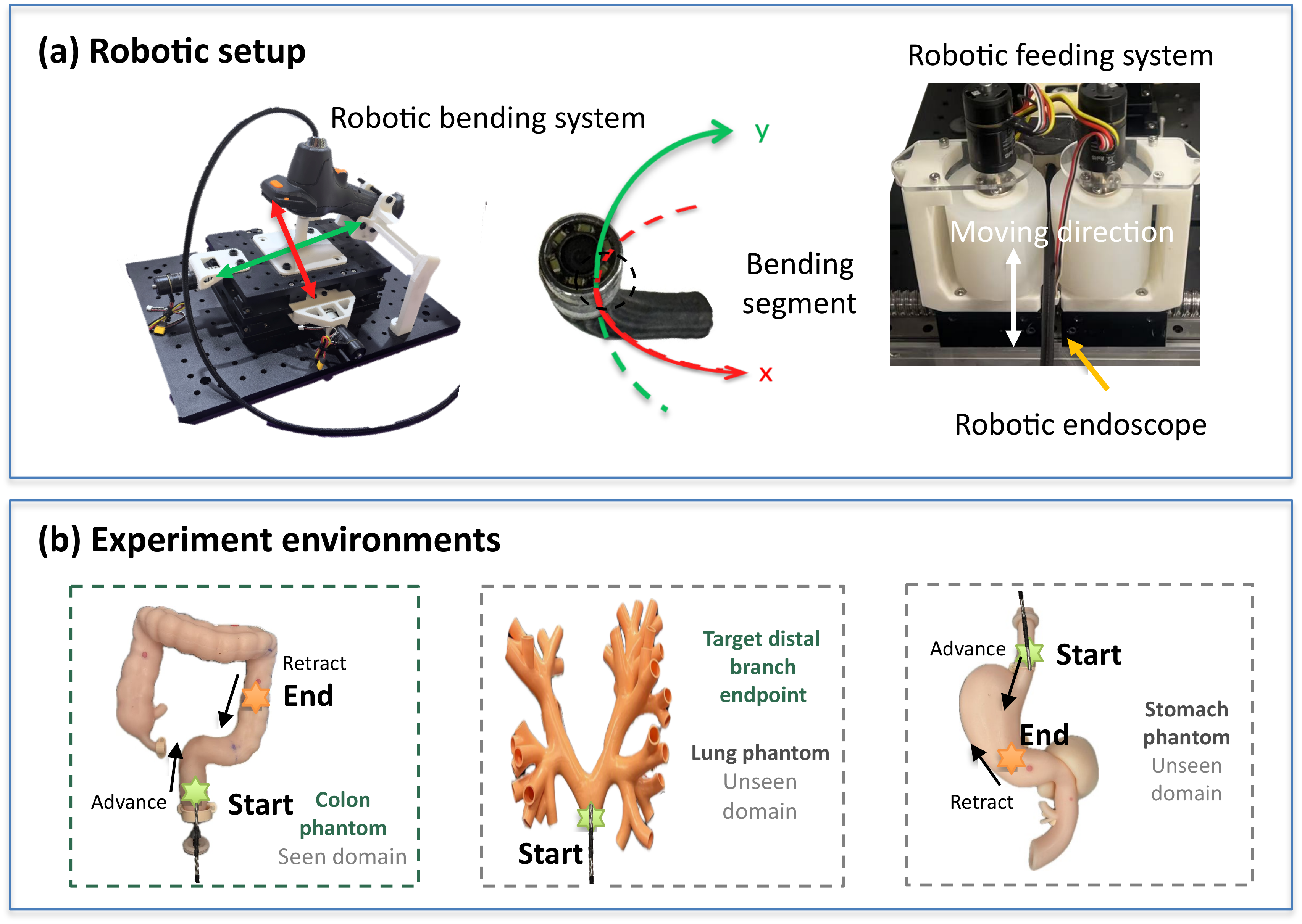}
\caption{Robotic platform and evaluated domains. (a) The setup combines a
motorized feeding unit for longitudinal advance/retraction with two-axis distal
bending, matching the policy's three action channels. (b) Bidirectional trials
use the colon phantom as the seen domain and lung and stomach phantoms as unseen
domains; stars mark the prescribed start and end regions, with lung trials
terminating at any target distal branch endpoint.}
\label{fig:robotic_sys_exp_setup}
\end{figure*}

\section{Related Work}
\label{sec:related}

\subsection{Autonomous Endoscopic Navigation and Bidirectional Motion}

Clinical colonoscopy separates insertion from careful withdrawal
inspection~\cite{barclay2006withdrawal}, and complete upper-gastrointestinal
examination requires systematic antegrade and retroflexed
views~\cite{emura2020photodocumentation}. A systematic review documents the
broader progression of autonomous navigation in intraluminal and endovascular
robotics~\cite{pore2023review}. Robotic work has reproduced
insertion--withdrawal structure for magnetic capsule
navigation~\cite{xu2022bidirectional}. Within flexible endoscopy, force sensing
and heuristic search support magnetic colonoscope
navigation~\cite{huang2021magnetic}; monocular depth, shape sensing, and
model-predictive control support geometry-aware steering~\cite{lu2023flexible};
and spatially aware monocular vision adapts motion to lumen
appearance~\cite{yang2023lumen}. A supervised colonoscopy system further
predicts steering targets and collision probability and can invoke withdrawal
through control logic~\cite{hwang2026colonoscopy}. Beyond endoscopy,
supervised autonomous soft-tissue surgery and laparoscopic intestinal
anastomosis demonstrate the value of tightly scoped task autonomy under
controlled conditions~\cite{shademan2016star,saeidi2022star}; the IDEAL
framework provides a staged route for evaluating such surgical-robotic systems
toward clinical translation~\cite{vasey2024ideal}. These methods motivate
bounded autonomy, while our work asks how one endoscopic policy should switch
axial direction when external procedural intent changes before visual geometry
does.

\subsection{Language-Conditioned VLA for Robotic and Endoscopic Control}

BC-Z, CLIPort, CALVIN, and VIMA study language- or multimodal-prompted robot
learning across task generalization, spatial manipulation, and long-horizon
execution~\cite{jang2022bcz,shridhar2022cliport,mees2022calvin,jiang2023vima}.
PaLM-E and Gato integrate embodied observations into generalist sequence
models~\cite{driess2023palme,reed2022gato}. RT-1, RT-2, and OpenVLA scale
language-conditioned control, while Open X-Embodiment and Octo broaden
cross-robot data and policy transfer~\cite{brohan2023rt1,zitkovich2023rt2,
kim2025openvla,oneill2024openx,ghosh2024octo}. R3M and DINOv2 provide
complementary evidence that broad visual pretraining can improve reusable robot
representations~\cite{nair2023r3m,oquab2024dinov2}. PaliGemma~2 and Qwen3-VL
provide transferable multimodal backbones~\cite{steiner2024paligemma2,
bai2025qwen3vl}, and GR00T N1 couples a VLM to a high-frequency action
model~\cite{bjorck2025groot}.

Within medical robotics, EndoVLA addresses prompt-conditioned tracking,
BiliVLA combines scene-aware navigation with reinforcement learning, and
EndoWAM introduces grounded world--action modeling for endoscopic
navigation~\cite{ng2025endovla,lin2026bilivla,lin2026endowam}.
Open-H-Embodiment complements these model-centric efforts with a large-scale
medical-robotics dataset initiative~\cite{openh2026embodiment}. Their
instructions organize tasks or visually recognized stages. Our instruction has
a narrower causal role: it selects between opposing axial modes under the same
observation, while visual observations condition geometry-dependent bending
within the selected axial mode.

\subsection{Generative Action Modeling and Intent-Conditioned Control}

Implicit Behavioral Cloning shows why explicit regression struggles with
multi-valued actions~\cite{florence2022ibc}. Behavior Transformers and
Diffusion Policy represent multimodal continuous
behavior~\cite{shafiullah2022bet,chi2023diffusion}; ACT shortens the effective
decision horizon through action chunking~\cite{zhao2023act}; and flow matching
and rectified flow learn continuous transport from noise to
actions~\cite{lipman2023flow,liu2023rectified}. $\pi_0$ conditions a flow-based
action expert on a vision-language prefix~\cite{black2025pi0}. These models can
represent several plausible trajectories, but multimodality alone does not
identify which externally requested phase is active. Our policy includes the
instruction in the visual-language context and uses a variational trajectory
latent to stochastically condition continuous action generation.

\section{Problem Formulation}
\label{sec:problem}

Let $\mathbf{o}_t=(I_t,\mathbf{s}_t)$ contain an RGB endoscopic frame and the
previous normalized action state. Let $m_t\in\{N,R\}$ denote normal navigation or urgent retraction, and let
$\tilde{\mathbf{A}}_t\in\mathbb{R}^{H\times 3}$ denote the normalized target
action chunk for longitudinal motion, up/down bending, and left/right bending. The implementation uses
\begin{equation}
 a_{\mathrm{fwd}}<0:\ \text{advance}, \qquad
 a_{\mathrm{fwd}}>0:\ \text{retract}.
 \label{eq:sign}
\end{equation}

Suppose an observation neighborhood occurs in both modes. Normal demonstrations
have typical longitudinal action $-v_N$ and urgent demonstrations have $+v_R$,
where $v_N,v_R>0$. If $q=\Pr(m=R\mid\mathbf{o})$, the population optimum of a
squared-error predictor is
\begin{equation}
 a^*_{\mathrm{MSE}}(\mathbf{o})
 =\mathbb{E}[a_{\mathrm{fwd}}\mid\mathbf{o}]
 =qv_R-(1-q)v_N.
 \label{eq:conditionalmean}
\end{equation}
Class imbalance can bias the regression toward one mode, whereas balanced
opposing modes can produce averaging or an unhelpful hold. In either case, the
core problem is conditional non-identifiability: $\mathbf{o}$ alone does not
identify the intended mode. A generative model can learn
$p(\mathbf{A}\mid\mathbf{o})$, but sampling still leaves the intended component
undefined. We therefore condition on a language variable
\begin{align}
 p_\theta(\mathbf{A}_t\mid I_t,\mathbf{s}_t,\ell_t),\quad
 \ell_t\in\{&\text{forward navigation},\nonumber\\[-1mm]
             &\text{urgent retraction}\}.
 \label{eq:conditionalpolicy}
\end{align}
These are the two canonical deployment anchors; semantically equivalent
surface forms used during training are described in
\secref{sec:data_language}.

The evaluation holds image and previous action fixed while switching the mode
condition, compares \methodname{} and \rfwithoutlatent{} on the same frozen split, and
measures closed-loop navigation and retraction on the seen colon phantom and
unseen lung and stomach phantoms. ModeFlag-LCRF separately tests a literal
one-bit interface once the upstream mode is known.

\begin{figure*}[!t]
\centering
\includegraphics[width=0.98\textwidth]{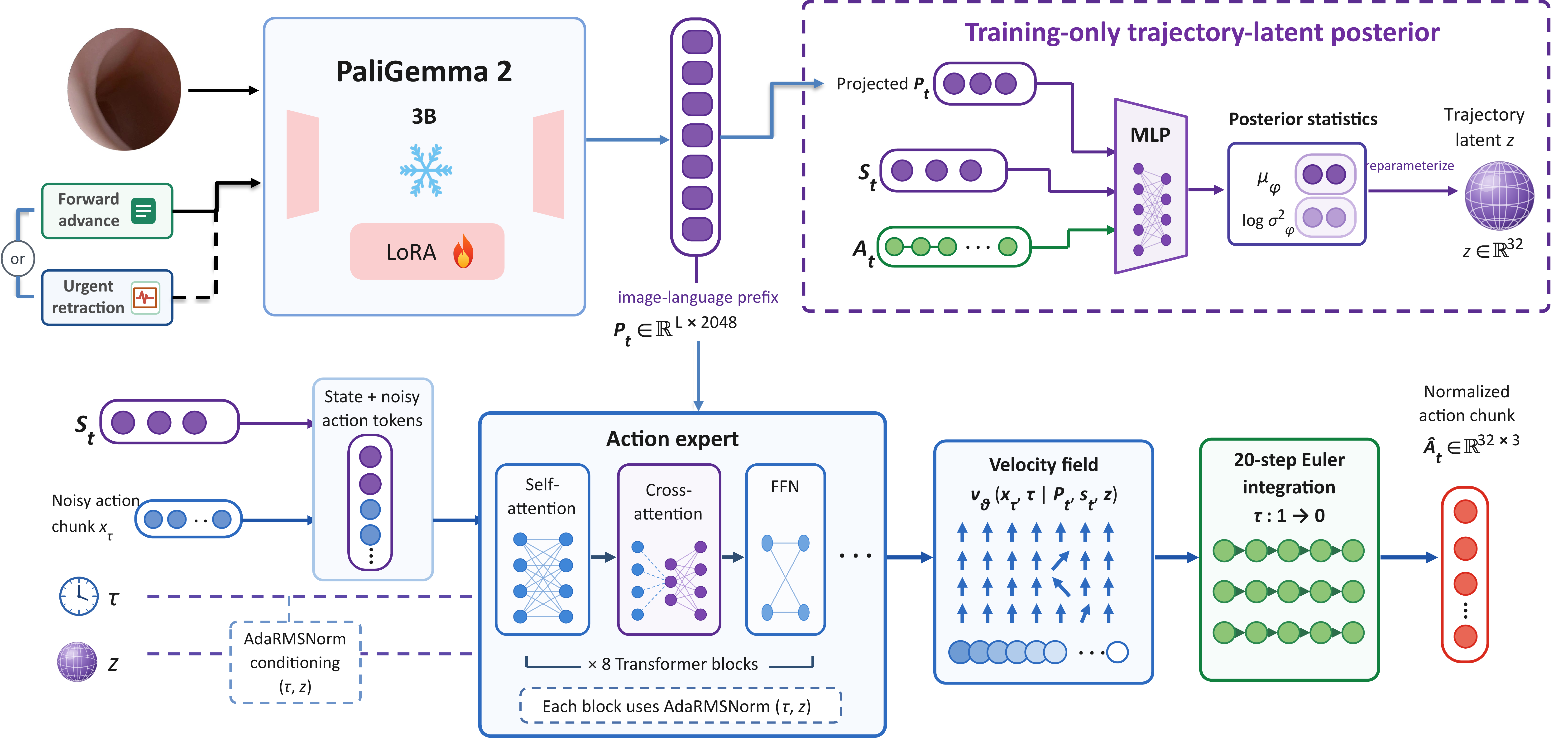}
\caption{EndoLIFT model architecture. The RGB frame and canonical instruction
are encoded by a LoRA-adapted PaliGemma~2 into prefix $\mathbf P_t$. The
previous executed action $\mathbf s_t$, noisy chunk $\mathbf x_\tau$, flow time
$\tau$, and trajectory latent $\mathbf z$ condition an eight-block Transformer
action expert, whose velocity field is integrated for 20 Euler steps to yield a
normalized $32{\times}3$ action chunk. The dashed upper branch is used only in
training: a posterior pools $\mathbf P_t$, $\mathbf s_t$, and the target chunk
$\mathbf A_t$ to parameterize $q_\phi(\mathbf z\mid\mathbf P_t,\mathbf s_t,
\mathbf A_t)$; deployment instead samples $\mathbf z\sim\mathcal N(0,\mathbf I)$.}
\label{fig:model}
\end{figure*}

\section{Method}
\label{sec:method}

\subsection{System Decomposition and Control Interface}

\figref{fig:overview}(b) separates mode selection from learned visual control.
Physiological measurements and speech are processed by an external trigger
supervisor; they are not concatenated with the image or passed to the neural
policy. The supervisor selects one of two fixed instructions, after which the
same VLA produces geometry-dependent motion.

The real-robot program uses a $480{\times}480$ RGB camera, a 20-Hz host and
motor-control loop, asynchronous VLA inference, a serial motor bridge, and four
motor targets. The three-dimensional policy action is mapped to paired axial
drive motors M1/M2, left/right bending motor M3, and up/down bending motor M4.
The physical platform and evaluated phantom domains are shown in
\figref{fig:robotic_sys_exp_setup}.

\subsection{Image-Language Prefix and Action State}

We use \texttt{google/paligemma2-3b-pt-224} with LoRA
adaptation~\cite{steiner2024paligemma2,hu2021lora}. The prompt combines the
current image, active instruction, action-channel definition, and a request for
the next normalized action chunk. The final PaliGemma hidden sequence
$\mathbf{P}_t\in\mathbb{R}^{L\times d_p}$ is retained as the visual-language
prefix. The action expert additionally receives the previous-action state
$\mathbf{s}_t$.

\subsection{Action Space}

The VLA policy predicts a three-dimensional continuous action
\begin{equation}
\mathbf{a}_t =
\left[
a_{\mathrm{fwd}},
a_{\mathrm{ud}},
a_{\mathrm{lr}}
\right]^{\top},
\end{equation}
where $a_{\mathrm{fwd}}$ controls longitudinal endoscope motion,
$a_{\mathrm{ud}}$ controls distal up--down bending, and
$a_{\mathrm{lr}}$ controls distal left--right bending.
Under the adopted sign convention, negative $a_{\mathrm{fwd}}$ denotes
forward advancement and positive $a_{\mathrm{fwd}}$ denotes retraction,
while the signs of $a_{\mathrm{ud}}$ and $a_{\mathrm{lr}}$ determine the
corresponding bending directions.

The three action channels are normalized using per-axis scales of
$[60,150,150]$ for longitudinal motion, up/down bending, and left/right
bending, respectively. The normalized action chunk is converted back to motor
command units before execution.
At deployment, the controller executes at most eight actions from each 32-step
prediction before acquiring the latest image and replanning. An instruction
change invalidates the active action chunk and triggers replanning under the
updated mode.

\subsection{Latent-Conditioned Rectified-Flow Expert}

The expert predicts a horizon $H=32$ with action dimension three. It projects
the prefix, state token, noisy action tokens, and learned chunk positions into a
width-1024 transformer with eight blocks and 16 attention heads. Each block uses
self-attention over state/action tokens, cross-attention to the VLM prefix, and
an FFN; adaptive RMS normalization is conditioned jointly on flow time and a
32-D \emph{\latentname} (\VTL). A posterior over the full target action chunk is
used during training, while samples from the standard-normal prior are used at
inference.
\figref{fig:model} traces the exact training and inference paths.

For target $\mathbf{A}$, noise
$\bm{\epsilon}\sim\mathcal{N}(0,\mathbf{I})$, and
$\tau\sim\mathcal{U}(0,1)$, the implementation constructs
\begin{equation}
 \mathbf{x}_\tau=(1-\tau)\mathbf{A}+\tau\bm{\epsilon},\qquad
 \mathbf{u}=\bm{\epsilon}-\mathbf{A}.
 \label{eq:flowpath}
\end{equation}
A posterior network pools the prefix, previous action, and target chunk:
\begin{equation}
 q_\phi(\mathbf{z}\mid\mathbf{P}_t,\mathbf{s}_t,\mathbf{A})
 =\mathcal{N}\!\left(\bm{\mu}_\phi,\operatorname{diag}
  (\bm{\sigma}_\phi^2)\right), \quad \mathbf{z}\in\mathbb{R}^{32}.
 \label{eq:posterior}
\end{equation}
The reparameterized trajectory-latent code is added to the time conditioning.
Writing $i$ for a batch element and $j$ for a latent dimension, the implemented
loss is
\begin{align}
 \mathcal L_{\mathrm{flow}}
 &=\mathbb E\!\left[\left\|
 v_\theta(\mathbf P_t,\mathbf s_t,\mathbf x_\tau,\tau,\mathbf z)
 -\mathbf u\right\|_2^2\right], \nonumber\\
 K_{ij}
 &=-\tfrac12\!\left(1+\log\sigma_{ij}^2-\mu_{ij}^2
 -\exp(\log\sigma_{ij}^2)\right), \nonumber\\
 \mathcal L_{\mathrm{KL}}^{\mathrm{FB}}
 &=\frac{1}{B}\sum_{i=1}^{B}\sum_{j=1}^{32}\max(K_{ij},0.5), \nonumber\\
 \mathcal L
 &=\mathcal L_{\mathrm{flow}}+\beta(s)\mathcal L_{\mathrm{KL}}^{\mathrm{FB}},
 \quad
 \beta(s)=10^{-3}\min\!\left(1,\frac{s}{500}\right).
 \label{eq:loss}
\end{align}
Thus the 0.5-nat free-bits floor is applied per latent dimension before
summation, and the KL weight increases linearly over 500 optimizer steps to its
final value of $10^{-3}$. At inference,
$\mathbf{z}\sim\mathcal{N}(0,\mathbf{I})$, a noise chunk is initialized at
$\tau=1$, and 20 Euler steps integrate to $\tau=0$. The matched ablation without
latent conditioning sets \texttt{latent\_dim=0}. In both variants, the language
instruction remains part of the PaliGemma~2 context.

\subsection{Architecture-Controlled ModeFlag Reference}
\label{sec:modeflag_method}

ModeFlag-LCRF is an architecture-controlled interface reference for \methodname{}. It retains the
PaliGemma~2 image path, previous-action token, 32-D trajectory latent, and the
same width-1024, eight-block rectified-flow action expert, but deletes the
instruction from the prompt. A constant neutral prompt is used for both modes,
and the implementation rejects any sample containing nonempty instruction
text. The literal flag $b\in\{0,1\}$ (advance/retract) is mapped by a learned
embedding and MLP to one prefix token and an adaptive-normalization condition.
Thus the 3B backbone is still used for vision, but it no longer encodes the
one-bit mode through language.

ModeFlag-LCRF is initialized from the \methodname{} checkpoint and fine-tuned
for two epochs using frozen-teacher action chunks from 29,672 colon frames.
During this stage, the LoRA parameters remain fixed, while the mode encoder and
action expert are optimized.

\subsection{External Trigger Supervisor and Instruction Switching}

Physiological measurements and speech features are not VLA inputs. A threshold monitor and an
asymmetric two-stage speech path generate discrete events. Normal operation
activates \emph{forward navigation}; an urgent physiological or speech event
activates \emph{urgent retraction}. Resumption is intended to require
normalized physiology and a high-confidence KEEP\_FORWARD event inside a finite
confirmation window. This separation keeps event detection and instruction selection outside the learned control policy.

\section{Experimental Protocol}
\label{sec:experiments}

\subsection{Robotic Platform and Phantom Domains}

The platform in \figref{fig:robotic_sys_exp_setup}(a) couples a robotic feeding
system to a two-axis bending mechanism. A $480{\times}480$ tip-camera stream is
processed asynchronously while the host and motor interface run at a configured
20~Hz. Colon is the seen phantom domain; lung and stomach are held out as unseen
phantom geometries. The start/end markers and nominal advance--retract paths in
\figref{fig:robotic_sys_exp_setup}(b) define the closed-loop tasks. The phantoms
are tested separately and are shown together only to summarize the protocol.

\subsection{Tasks, Instructions, and Data Composition}
\label{sec:data_language}

The learning problem contains two axial modes: forward motion during navigation
and retraction after an urgent trigger. The 44,942-frame base imitation dataset
contains four instruction strings (\tabref{tab:instructions}): three surface
forms correspond to forward motion and one corresponds to retraction. The
external trigger supervisor uses \emph{forward navigation} and
\emph{urgent retraction} as the two canonical deployment instructions.
The \emph{forward navigation} anchor is introduced during the subsequent
instruction-conditioning stage rather than appearing in the base imitation
dataset.

\begin{table}[t]
\centering
\caption{Instruction composition of the base imitation dataset. The four
surface forms correspond to two axial control modes; the canonical deployment
instructions are defined in the text.}
\label{tab:instructions}
\small
\begin{tabular}{@{}>{\raggedright\arraybackslash}p{0.55\columnwidth}
                @{\hspace{4pt}}>{\centering\arraybackslash}p{0.17\columnwidth}
                @{\hspace{4pt}}>{\raggedleft\arraybackslash}p{0.19\columnwidth}@{}}
\toprule
Base-table instruction & Mode & Frames \\
\midrule
\emph{failed recovery navigation} & forward & 8,344 \\
\emph{normal navigation} & forward & 6,850 \\
\emph{successful recovery navigation} & forward & 1,278 \\
\emph{urgent retraction} & retract & 28,470 \\
\midrule
\textbf{Total} & & \textbf{44,942} \\
\bottomrule
\end{tabular}
\end{table}

Following base imitation training, the policy undergoes 2,000
instruction-conditioning updates using 1,200 sampled frames, a pool of
40 forward and 40 retraction paraphrases with on-the-fly perturbations, and a
20\% canonical-anchor mixture. Instruction generalization is evaluated on
held-out surface forms after this conditioning stage.

\begin{figure*}[!t]
\centering
\includegraphics[width=0.94\textwidth]{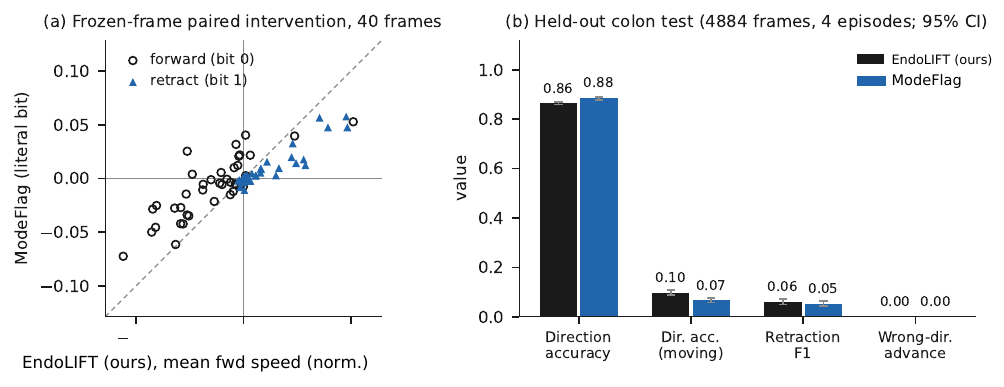}
\caption{Architecture-controlled ModeFlag reference comparison. (a) Mean normalized longitudinal
output for EndoLIFT's single canonical anchor (horizontal) and
ModeFlag-LCRF's literal bit (vertical) on the same 40 frames and noise seeds.
(b) Additional held-out colon diagnostic on 4,884 frames from four episodes;
bars show 95\% bootstrap intervals. Overall accuracy, moving-only accuracy,
retraction F1, and wrong-direction advance are reported together because hold
steps are prevalent.}
\label{fig:m1_modeflag}
\end{figure*}

\subsection{Baselines and Ablations}

We use five compact model names throughout the text, tables, and figures.
\dinobaseline{} is the no-language baseline: DINOv2 supplies a
self-supervised visual representation~\cite{oquab2024dinov2}, which is
concatenated with the previous-action state. In the model names, MLP denotes a
direct regression action head. \qwenmlp{} instead uses the Qwen3-VL
image--language representation~\cite{bai2025qwen3vl} with the same MLP head,
whereas \qwengroot{} uses a generative flow action head based on
GR00T~\cite{bjorck2025groot}. \rfwithoutlatent{} is the matched
PaliGemma~2 rectified-flow ablation, and \methodname{} adds the 32-D trajectory
latent. Thus, the matched comparison changes only trajectory-latent
conditioning, while the other rows test alternative visual and action-head
designs.
All five policies are reported on the same 185-episode table and frozen
episode-level split. We additionally report
ModeFlag-LCRF (\secref{sec:modeflag_method}) and two explicit language-to-bit
front ends: a keyword router and a character TF--IDF router trained from the
five policy-visible instruction anchors. The routers emit a bit and select the
corresponding single canonical string for the frozen \methodname{} policy.

\subsection{Mode-Interface and Instruction-Generalization Evaluation}
\label{sec:m1_protocol}

The primary M1 intervention freezes 40 images (20 advance-scene and 20
retraction-scene frames), previous-action state, and per-frame flow-noise seed.
For each frame, ModeFlag-LCRF receives both literal bits and \methodname{}
receives the two single canonical anchors. Intent-following accuracy (IFA) is
the fraction for which the sign of mean longitudinal action over the 32-step
chunk matches the requested intent. Canonical paired flip is the fraction of
frames with negative output under the canonical forward anchor and positive
output under the canonical retract anchor; separation is the mean
retract-minus-forward output gap.

Instruction generalization uses the same frozen frame pool and 44 held-out
commands: 12 English paraphrases, eight Chinese paraphrases, eight unseen oral
commands, ten noisy/ASR variants, and six long commands containing distractors.
We report category IFA and the example-weighted All-unseen aggregate. An entry
is marked N/A when a method has no language interface; N/A is never replaced by
an artificial zero. The native 1-bit baseline evaluates control after intent is
known, whereas the router pipelines include the conversion from free-form
language to that bit.

\subsection{Policy-Level Evaluation}

We use an episode-level 90/5/5 train/validation/test split with seed 42,
comprising 166 training episodes, nine validation episodes, and ten test
episodes, for 185 episodes in total. The test split contains 150,351 valid
frame--chunk-step pairs after application of the horizon mask, including
16,831 pairs from forward-mode frames and 10,345 true-retraction target steps.
No episode is shared across the training, validation, and test partitions.

Navigation direction accuracy is defined as three-way
advance/hold/retract classification accuracy on forward-mode frames. Retraction precision,
recall, and F1 are computed on the retract decision; reporting all three avoids
rewarding a policy that predicts retract almost everywhere. We define the
\emph{critical advance rate} as the fraction of true-retract target steps
predicted as advance. Forward and bending MAE are retained as conventional
regression diagnostics. Point estimates are accompanied by 95\% episode-
bootstrap intervals.

Conditioning is evaluated through three interventions: (i) swap the instruction
while holding image and previous action fixed; (ii) replace text with the
architecture-controlled ModeFlag reference; and (iii) shuffle the image while retaining mode
and state.

\subsection{Closed-Loop Phantom Evaluation}

The closed-loop pilot designates colon as the seen phantom domain and lung and
stomach as unseen phantom domains. Each method-condition cell contains ten
trials: 20 seen-domain trials and 40 unseen-domain trials per method, yielding
300 reported trials across five methods. Navigation and urgent retraction are
reported separately within each domain. Success rates are accompanied by exact
95\% Clopper--Pearson binomial intervals~\cite{clopper1934interval}, and
\methodname{} is compared with \rfwithoutlatent{} using two-sided
Fisher exact tests~\cite{agresti1992exact}. The accompanying transition graph
summarizes 0.5-s decision windows from real-robot closed-loop recordings.
Transition graphs are episode-level summaries; the fixed-input instruction
intervention provides the controlled mode-switch comparison.

\subsection{Ex-Vivo Porcine-Trachea Evaluation}
\label{sec:exvivo_protocol}

We further evaluate the frozen \methodname{} policy in ten triggered-retraction
trials in a porcine trachea. Each trial begins in forward-navigation mode; a
human-triggered event then causes the external trigger supervisor to switch the
instruction to urgent retraction.
We report trial success, the two trigger-to-action latencies, post-trigger axial
direction, residual bending activity, and human takeover. Trial success uses an
exact 95\% Clopper--Pearson interval; latencies are reported as
mean~$\pm$~standard deviation with the observed range.

\section{Results}
\label{sec:results}

\subsection{Literal 1-bit Baseline and Held-Out Instructions}

\begin{table}[t]
\centering
\caption{Architecture-controlled ModeFlag reference versus canonical-text intervention on 40 frozen frames.
Image, previous action, and per-frame flow-noise seed are identical. IFA and
canonical paired flip are percentages; separation is normalized longitudinal
action.}
\label{tab:m1_modeflag}
\small
\setlength{\tabcolsep}{3.2pt}
\begin{tabularx}{\columnwidth}{@{}Xccc@{}}
\toprule
Condition interface & IFA $\uparrow$ & \shortstack{Canonical\\paired flip} $\uparrow$ & Separation $\uparrow$ \\
\midrule
Literal 1-bit flag (ModeFlag-LCRF) & 68.8 & 37.5 & 0.0166 \\
Canonical text (EndoLIFT) & \textbf{85.0} & \textbf{70.0} & \textbf{0.0476} \\
\bottomrule
\end{tabularx}
\end{table}

On the paired 40-frame intervention, canonical text improves IFA by 16.2
percentage points and canonical paired flip by 32.5 points over the literal
bit, while the mean conditional separation is 2.86$\times$ larger
(\tabref{tab:m1_modeflag}). Under this training protocol, the architecture-controlled ModeFlag reference exhibits weaker
same-observation switching than the text-conditioned policy.

\begin{table*}[t]
\centering
\caption{Instruction-following accuracy (IFA, \%) on the frozen 40-frame pool.
``All unseen'' aggregates 44 held-out commands. Multi-instruction flip is
computed over the multi-string instruction pool used in this table; it is
distinct from the two-anchor canonical paired flip in
Table~\ref{tab:m1_modeflag} and the test-set instruction-swap flip in
Table~\ref{tab:offline_language}. The ModeFlag row uses literal bits and
therefore has no language-category or multi-instruction value (``--'' denotes
N/A). The \ourslabel{} canonical cell averages three seen canonical strings.}
\label{tab:m1_instruction_generalization}
\footnotesize
\setlength{\tabcolsep}{1.6pt}
\renewcommand{\arraystretch}{1.05}
\begin{tabularx}{\textwidth}{@{}>{\raggedright\arraybackslash}p{0.235\textwidth}*{8}{>{\centering\arraybackslash}X}@{}}
\toprule
Method & Seen canon. & EN para. & ZH para. & Oral & ASR/noise & Distr. & All unseen & \shortstack{Multi-instr.\\flip} \\
\midrule
\qwenmlp{} & 57.5 & 60.0 & 60.6 & 57.5 & 59.5 & 61.3 & 59.7 & 20.0 \\
\qwengroot{} & 74.2 & 62.9 & 62.2 & 66.6 & 59.5 & 65.8 & 63.1 & 30.0 \\
\rfwithoutlatent{} & 90.8 & 79.0 & 79.1 & 77.8 & 75.5 & 80.4 & 78.2 & 55.0 \\
Keyword router + canonical policy & 85.8 & 50.0 & 50.0 & 50.0 & 57.0 & 61.7 & 53.2 & 5.0 \\
TF--IDF router + canonical policy & 85.8 & 50.0 & 50.0 & 67.5 & 57.0 & 50.0 & 54.8 & 7.5 \\
ModeFlag-LCRF (literal bit; no text) & 68.8 & -- & -- & -- & -- & -- & -- & -- \\
\textbf{\ourslabel{}} & \textbf{90.0} & \textbf{84.2} & \textbf{83.4} & \textbf{85.3} & \textbf{77.8} & \textbf{84.6} & \textbf{82.8} & \textbf{55.0} \\
\bottomrule
\end{tabularx}
\end{table*}

Across 44 held-out surface forms, \methodname{} achieves 82.8\% IFA, compared
with 78.2\% for \rfwithoutlatent{}, 63.1\% for \qwengroot{},
and 59.7\% for \qwenmlp{}
(\tabref{tab:m1_instruction_generalization}). The keyword and TF--IDF
router--policy pipelines reach only 53.2\% and 54.8\% downstream IFA; their
router intent accuracies are 54.5\% and 56.8\%, respectively. ModeFlag-LCRF has
no entry in these language columns because it accepts only the mode bit. The
text-conditioned policy directly accepts the evaluated instruction variants,
whereas ModeFlag begins after the language-to-intent conversion.

\subsection{Language Selects the Axial Motion Mode Independently of VTL}

The test-set instruction-swap diagnostic in \tabref{tab:offline_language}
changes only the instruction while holding the image and previous action fixed.
Switching from \emph{forward navigation} to \emph{urgent retraction} shifts
\methodname{}'s longitudinal output by 3.484 raw motor units toward retraction;
68.1\% of paired predictions move in the requested direction and 45.2\% cross
the axial sign boundary. Because the visual and action-state inputs are
unchanged, this establishes that language changes the selected axial mode.
The matched \rfwithoutlatent{} ablation produces a larger instruction-induced
response in this diagnostic, with a shift of 6.014, a correct-sign fraction of
0.856, and a test-set instruction-swap flip of 0.644. Together with the stronger
directional and closed-loop results of \methodname{}, this dissociation shows
that instruction sensitivity and trajectory execution quality capture distinct
properties of the controller: language selects the requested axial mode, while
VTL conditioning improves execution of the resulting trajectory.

\begin{table}[t]
\centering
\caption{Test-set instruction-swap diagnostic with image and previous-action
state fixed. Shift is the axial-output change [95\% CI] in raw motor units; CSF
is correct-sign fraction and test-swap flip is the test-set instruction-swap
flip rate. These sensitivity diagnostics are interpreted separately from task-performance metrics.}
\label{tab:offline_language}
\scriptsize
\setlength{\tabcolsep}{1.8pt}
\begin{tabularx}{\columnwidth}{@{}>{\raggedright\arraybackslash}X>{\centering\arraybackslash}p{0.39\columnwidth}cc@{}}
\toprule
Method & Shift [95\% CI] & CSF & \shortstack{Test-swap\\flip} \\
\midrule
\qwenmlp{} & 1.615 [1.399, 2.683] & 0.979 & 0.376 \\
\qwengroot{} & 1.587 [1.234, 4.645] & 0.613 & 0.248 \\
\rfwithoutlatent{} & 6.014 [5.762, 7.983] & 0.856 & 0.644 \\
\ourslabel{} & 3.484 [3.234, 5.503] & 0.681 & 0.452 \\
\bottomrule
\end{tabularx}
\end{table}

\begin{figure}[t]
\centering
\includegraphics[width=\columnwidth]{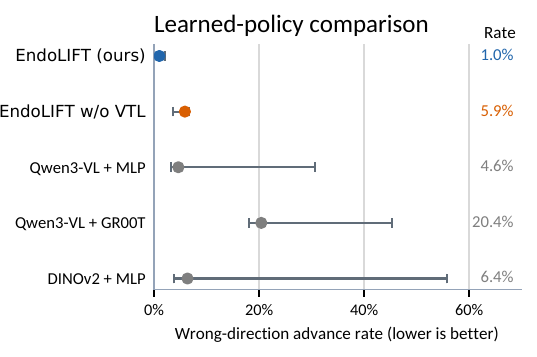}
\caption{Wrong-direction advance on held-out true-retraction targets under the
correct instruction; lower is better. Points indicate the observed rates, and
error bars denote 95\% confidence intervals. \rfwithoutlatent{} is the matched
ablation without trajectory-latent conditioning.}
\label{fig:offline_quant}
\end{figure}

\subsection{Visual Observations Condition Bending Within the Selected Mode}

Shuffling the image increases bending MAE from 13.742 to 14.148 overall and
from 32.927 to 34.227 on forward-navigation samples, demonstrating measurable
visual dependence of bending predictions.

\subsection{Latent Conditioning Improves Directional Correctness}

\begin{table*}[t]
\centering
\caption{Directional performance on the held-out test set. Navigation direction
accuracy and wrong-direction advance are reported with 95\% episode-bootstrap
confidence intervals. The last two columns report the difference between
\ourslabel{} and \rfwithoutlatent{}.}
\label{tab:offline_main}
\scriptsize
\setlength{\tabcolsep}{3.1pt}
\begin{tabularx}{\textwidth}{@{}Xcccc@{}}
\toprule
Method & Nav. direction $\uparrow$ & Wrong-direction advance $\downarrow$ &
$\Delta$ Nav. vs. matched & Error reduction vs. matched \\
\midrule
\dinobaseline{} & 0.395 [0.328, 0.580] & 0.064 [0.037, 0.557] & -- & -- \\
\qwenmlp{} & 0.392 [0.324, 0.576] & 0.046 [0.033, 0.307] & -- & -- \\
\qwengroot{} & 0.351 [0.288, 0.535] & 0.204 [0.181, 0.453] & -- & -- \\
\rfwithoutlatent{} & 0.453 [0.412, 0.589] & 0.059 [0.036, 0.067] & reference & reference \\
\textbf{\ourslabel{}} & \textbf{0.565 [0.534, 0.668]} & \textbf{0.010 [0.005, 0.021]} & \textbf{+11.1 pp} & \textbf{83\%} \\
\bottomrule
\end{tabularx}
\vspace{-3mm}
\end{table*}

Relative to \rfwithoutlatent{}, \methodname{} improves navigation-direction
accuracy by 11.1 percentage points
and reduces wrong-direction advance on retraction targets by 83\%
(\tabref{tab:offline_main} and \figref{fig:offline_quant}). The corresponding
95\% confidence intervals do not overlap. The two models share the same
image-language backbone, state input, action horizon, and flow objective,
differing only in trajectory-latent conditioning.
These task metrics improve despite the larger test-set instruction-swap response
of \rfwithoutlatent{} in \tabref{tab:offline_language}, showing that greater
mode sensitivity is not equivalent to better control. Together with the
representative action chunks below, the results indicate that VTL contributes
to directional correctness and coherent three-axis trajectory execution.

\subsection{Action Chunks Preserve Three-Axis Temporal Structure}

\begin{figure*}[!t]
\centering
\includegraphics[width=\textwidth]{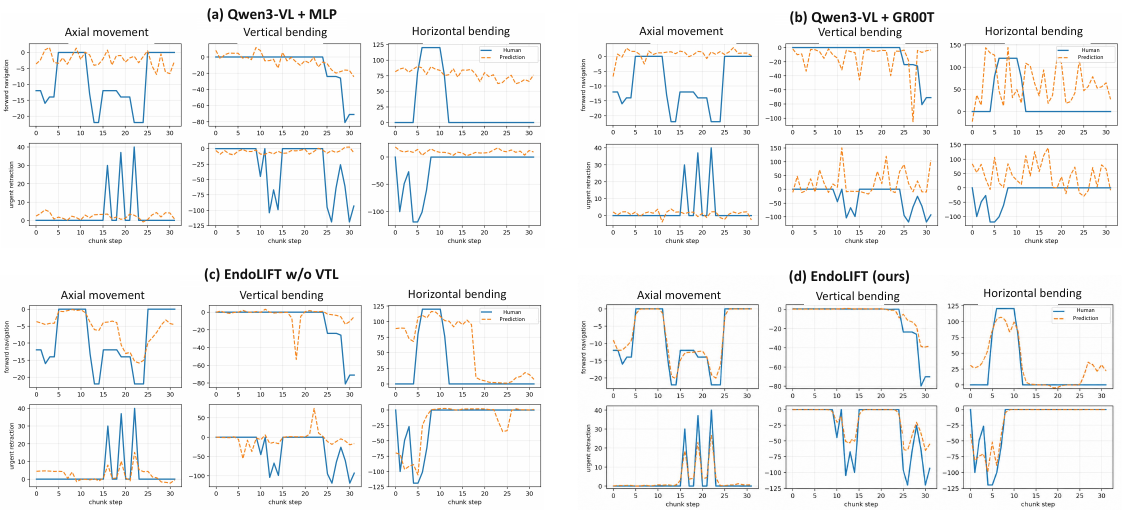}
\caption{Frozen-test action-chunk comparison for (a) \qwenmlp{},
(b) \qwengroot{}, (c) \rfwithoutlatent{}, and (d) \ourslabel{}. Within each
panel, the top and bottom rows show forward navigation and urgent retraction,
respectively; the columns show axial movement, vertical bending, and horizontal
bending. Blue solid curves are human targets, and orange dashed curves are
open-loop predictions.}
\label{fig:qualitative}
\vspace{-3mm}
\end{figure*}

\figref{fig:qualitative} provides a temporal view that complements the
aggregate comparison. During forward navigation, \methodname{} follows the
negative longitudinal plateaus and returns toward zero during inactive
intervals. \rfwithoutlatent{} underestimates early and late
axial segments, retains a negative bias after the final transition, and
prolongs the left/right bend beyond the target interval. \qwenmlp{}
largely collapses toward low-amplitude actions. In the representative example,
\qwengroot{} shows sign-inconsistent axial values and large spurious bending.

During urgent retraction, the proposed prediction aligns with the three
positive axial events and reproduces the timing of the dominant negative
bending intervals. \rfwithoutlatent{} is positively biased before the target
events and underestimates their peaks; \qwenmlp{} and \qwengroot{} either
suppress the events or oscillate on the bending axes.

\subsection{Instruction-Conditioned Replanning}

\begin{figure*}[!t]
\centering
\includegraphics[width=0.98\textwidth,trim=0 120bp 0 120bp,clip]{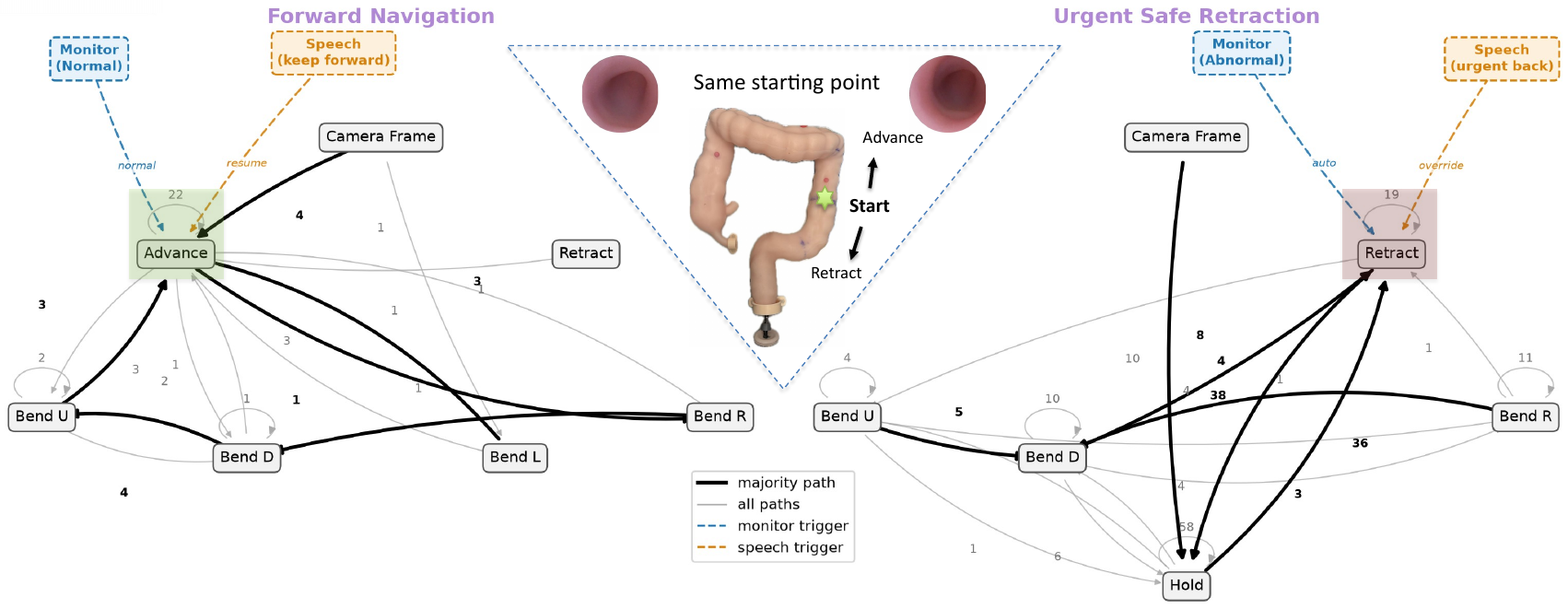}
\caption{Instruction-stratified validation transition structure for forward
navigation (left) and urgent retraction (right), with the common
bidirectional task illustrated centrally. Black arrows denote majority
transitions, gray arrows all observed transitions, and dashed blue/orange
arrows the monitor/speech interfaces.}
\label{fig:graph_offline}
\vspace{-4mm}
\end{figure*}

The validation transition graph in \figref{fig:graph_offline} is computed
separately for recorded forward-navigation and urgent-retraction episodes.
Under \emph{forward navigation}, dominant paths converge on Advance, with
bending actions returning to forward motion. Under
\emph{urgent retraction}, monitor and speech triggers enter the retraction
subgraph, and dominant transitions connect Hold and the bending states to
Retract. The two instruction groups exhibit different action-transition
patterns.

\begin{table}[t]
\centering
\caption{Closed-loop pilot on the seen colon phantom. Entries are success rate
in percent [exact 95\% CI]; $n=10$ per mode and $n=20$ overall.}
\label{tab:pilot_seen}
\scriptsize
\setlength{\tabcolsep}{1.2pt}
\begin{tabularx}{\columnwidth}{@{}>{\raggedright\arraybackslash}Xccc@{}}
\toprule
Method & Nav. & Retr. & Overall \\
\midrule
\dinobaseline{} & 0 [0,30.8] & 0 [0,30.8] & 0 [0,16.8] \\
\qwengroot{} & 0 [0,30.8] & 0 [0,30.8] & 0 [0,16.8] \\
\qwenmlp{} & 40 [12.2,73.8] & 0 [0,30.8] & 20 [5.7,43.7] \\
\rfwithoutlatent{} & 100 [69.2,100] & 40 [12.2,73.8] & 70 [45.7,88.1] \\
\textbf{\ourslabel{}} & \textbf{100 [69.2,100]} & \textbf{100 [69.2,100]} & \textbf{100 [83.2,100]} \\
\bottomrule
\end{tabularx}
\end{table}

\begin{table}[!t]
\centering
\caption{Quantitative ex-vivo porcine-trachea validation of \methodname{} over
ten human-triggered retraction trials.}
\label{tab:exvivo}
\scriptsize
\setlength{\tabcolsep}{3.0pt}
\renewcommand{\arraystretch}{1.05}
\begin{tabularx}{\columnwidth}{@{}>{\raggedright\arraybackslash}p{0.48\columnwidth}>{\raggedright\arraybackslash}X@{}}
\toprule
Metric & Result \\
\midrule
Trial success & 10/10; exact 95\% CI [69.2, 100]\% \\
Total duration / control steps & 162.9~s / 3,257 steps \\
Human trigger $\rightarrow$ instruction switch & $445\pm126$~ms (350--700) \\
Instruction switch $\rightarrow$ first Retract & $54\pm61$~ms (23--224); $\approx$1 step at 20~Hz \\
Retract among post-trigger axial-dominant steps & 1,127/1,127 = 100\% \\
Post-trigger bending activity & 5.2\% (36.4\% pre-trigger) \\
Human takeover & 0; \texttt{human\_takeover}=0 throughout \\
\bottomrule
\end{tabularx}
\end{table}

\subsection{Latent Conditioning Improves Bidirectional Success on the Seen Colon Phantom}

On the seen colon phantom, \methodname{} raises overall success by 30
percentage points (43\% relative) over \rfwithoutlatent{}
(\tabref{tab:pilot_seen}). Navigation performance is unchanged between the two
models, while urgent-retraction success increases by 60 percentage points.
An observed 10/10 rate has an exact 95\% CI of [69.2, 100]\%. Against
\rfwithoutlatent{}, the
retraction comparison is 10/10 versus 4/10 (two-sided Fisher exact
$p=0.0108$), and the overall comparison is 20/20 versus 14/20 ($p=0.0202$).
The gain is concentrated in reverse motion; navigation remains 10/10 for both
models. In \figref{fig:closed_loop_demo}(c), both policies advance before the
switch, but the \rfwithoutlatent{} rollout becomes stuck after initiating retraction.

\subsection{The Bidirectional Policy Transfers to Unseen Phantoms}

On the unseen lung and stomach phantoms, \methodname{} again improves
overall success by 30 percentage points (43\% relative) over
\rfwithoutlatent{} (\tabref{tab:pilot_unseen}). Both models retain the same navigation
success, whereas latent conditioning is associated with a 60-percentage-point
gain in retraction success across the two unseen geometries. Pooling the two
unseen retraction conditions gives 20/20 for \methodname{} versus 8/20 for
\rfwithoutlatent{} (two-sided Fisher exact $p=4.51\times10^{-5}$); the corresponding
overall comparison is 40/40 versus 28/40 ($p=1.85\times10^{-4}$). The proposed
policy reaches 10/10 in all four unseen conditions, while the matched policy
reaches 3/10 and 5/10 in lung and stomach retraction. Advance and retract
sequences for all three geometries are shown in
\figref{fig:closed_loop_demo}(b).
The 0.565 offline navigation-direction accuracy in
\tabref{tab:offline_main} is a frame-level three-way advance/hold/retract
measure, not episode-level success. Hold and transition ambiguity depress this
local metric, whereas receding-horizon replanning can correct isolated errors;
hence it does not conflict with 10/10 navigation success in each phantom.

\begin{table*}[!t]
\centering
\caption{Closed-loop transfer pilot on the unseen lung and stomach phantoms
($n=10$ per condition; $n=40$ overall). Each cell reports success rate in
percent [exact 95\% CI].}
\label{tab:pilot_unseen}
\scriptsize
\setlength{\tabcolsep}{1.0pt}
\renewcommand{\arraystretch}{1.05}
\begin{tabular}{@{}>{\raggedright\arraybackslash}p{0.245\textwidth}@{\hspace{2pt}}ccccc@{}}
\toprule
Method & \shortstack{Nav.-Lung\\SR [95\% CI]} &
\shortstack{Nav.-Stomach\\SR [95\% CI]} &
\shortstack{Ret.-Lung\\SR [95\% CI]} &
\shortstack{Ret.-Stomach\\SR [95\% CI]} &
\shortstack{Overall\\SR [95\% CI]} \\
\midrule
\dinobaseline{} & 0 [0, 30.8] & 0 [0, 30.8] & 0 [0, 30.8] & 0 [0, 30.8] & 0 [0, 8.8] \\
\qwengroot{} & 0 [0, 30.8] & 0 [0, 30.8] & 0 [0, 30.8] & 0 [0, 30.8] & 0 [0, 8.8] \\
\qwenmlp{} & 80 [44.4, 97.5] & 50 [18.7, 81.3] & 0 [0, 30.8] & 0 [0, 30.8] & 32.5 [18.6, 49.1] \\
\rfwithoutlatent{} & 100 [69.2, 100] & 100 [69.2, 100] & 30 [6.7, 65.2] & 50 [18.7, 81.3] & 70 [53.5, 83.4] \\
\textbf{\ourslabel{}} & \textbf{100 [69.2, 100]} & \textbf{100 [69.2, 100]} & \textbf{100 [69.2, 100]} & \textbf{100 [69.2, 100]} & \textbf{100 [91.2, 100]} \\
\bottomrule
\end{tabular}
\end{table*}

\subsection{Demonstrations Resolve Intent Aliasing and Execute Triggered Reversal}

In \figref{fig:closed_loop_demo}(a), the robot advances before the trigger and
retracts after the instruction changes, although the endoscopic view at the
switch is nearly unchanged. Camera feedback continues to condition bending
during the reversed axial motion.

\subsection{Cross-Domain Replanning Preserves Visual Steering}

\begin{figure*}[!t]
\centering
\includegraphics[width=\textwidth,trim=4bp 174bp 4bp 165bp,clip]{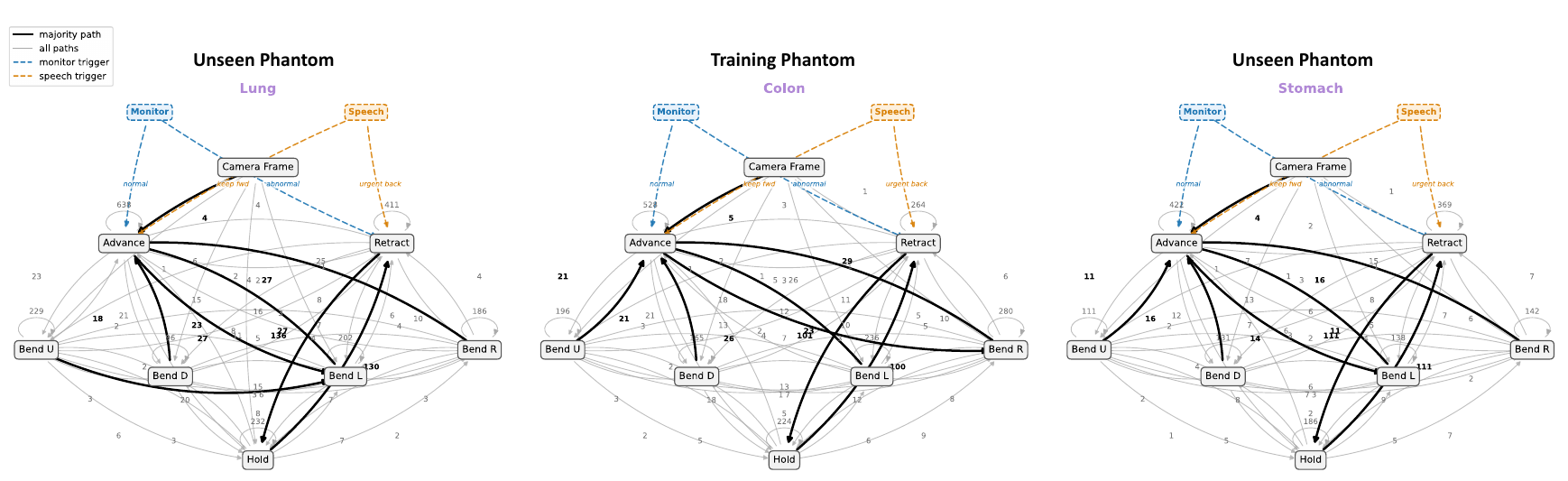}
\caption{Domain-stratified closed-loop action transitions for the unseen lung,
seen colon, and unseen stomach phantoms (left to right). Edges aggregate
0.5-s executed-action windows; black arrows show majority transitions, gray
arrows all observed transitions, and dashed arrows the monitor/speech
interfaces.}
\label{fig:graph_online}
\end{figure*}

The graphs in \figref{fig:graph_online} aggregate 0.5-s executed-action windows
within each phantom. Across all three domains, camera-conditioned bending and
hold states connect to axial Advance or Retract states, while monitor and
speech routes provide the external mode interface. Bending-to-axial transitions
occur in both modes, consistent with continued visual steering during
retraction.

\subsection{Closed-Loop Demonstrations and Ex-Vivo Validation}

\begin{figure*}[!t]
\centering
\includegraphics[width=0.98\textwidth]{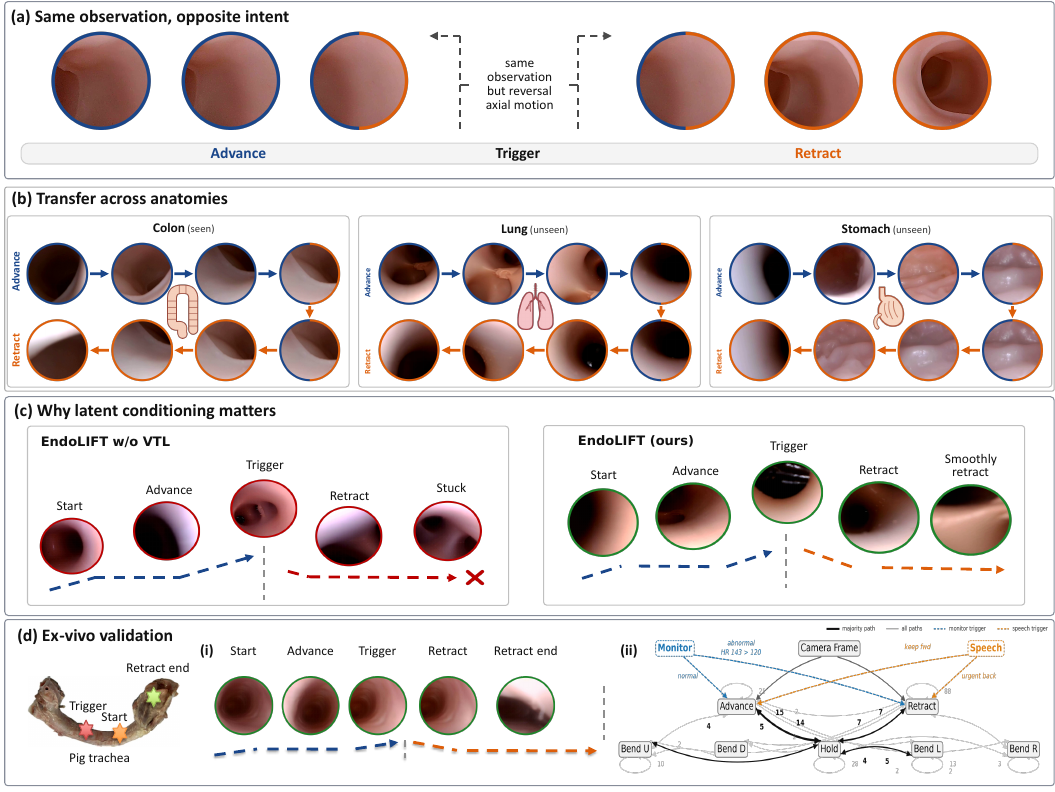}
\caption{Closed-loop bidirectional behavior and transfer. (a) The robot
advances until an external trigger switches the instruction, then retracts from
a nearly unchanged observation. (b) Advance and retract sequences in the seen
colon and unseen lung and stomach phantoms. (c) Trigger-centered rollouts for
\rfwithoutlatent{} and EndoLIFT; the former becomes stuck after initiating
retraction, whereas the latter retracts smoothly. (d) Ex-vivo validation in a
porcine trachea, including a representative trigger-centered sequence and
the corresponding action-transition graph.}
\label{fig:closed_loop_demo}
\end{figure*}

The final summary in \figref{fig:closed_loop_demo} links the same-observation
intent switch to closed-loop transfer and shows a representative ex-vivo
porcine-trachea trial. Across all ten ex-vivo trials, \methodname{} succeeds in
10/10 cases (exact 95\% CI [69.2, 100]\%) over 162.9~s and 3,257 control steps
(\tabref{tab:exvivo}). A human trigger reaches the instruction
switch in $445\pm126$~ms (range 350--700~ms), and the first retract command
follows the switch in $54\pm61$~ms (23--224~ms), approximately one control step
at 20~Hz. All 1,127 post-trigger axial-dominant steps are Retract. Bending
remains active in 5.2\% of post-trigger steps, down from 36.4\% before the
trigger, and no trial requires human takeover.

\section{Failure Case Study}
\label{sec:failure}

\paragraph{Retraction attenuation in \rfwithoutlatent{}}
\rfwithoutlatent{} retains forward navigation but shows a
60-percentage-point retraction deficit on both the seen and unseen evaluations.
Its selected urgent-retraction chunk also underestimates the axial peaks and
introduces positive bias before target events (\figref{fig:qualitative}(c)).

\paragraph{Unstable generic action generation}
In the selected \qwengroot{} example, the axial prediction changes sign and
the bending channels oscillate despite smooth targets
(\figref{fig:qualitative}(b)). The same baseline shows a 100-percentage-point
overall success deficit relative to \methodname{} in the reported closed-loop
conditions. The sign changes and oscillations localize this deficit to an
unstable, geometry-incompatible action sequence.

\section{Discussion}
\label{sec:discussion}

\paragraph{Clinical implication}
Bidirectional endoscopy requires both a procedural phase and geometry-aware
steering. A canonical instruction selects axial advance or retraction, and the
image-conditioned policy steers within that phase. The tested interface covers
urgent operator-requested reversal and extends naturally to
insertion--inspection workflows in which phase changes originate outside the
camera view.

\paragraph{What language adds beyond one bit}
A reliable upstream forward/retract bit provides a compact discrete control
interface, and the architecture-controlled ModeFlag reference makes that option explicit. When the external interface is speech or text, \methodname{} directly accepts
the tested linguistic variants while retaining visual steering within the same
policy. In the paired intervention, the text-conditioned policy produces a
larger mode-separation response than ModeFlag-LCRF.

\paragraph{Role of the variational trajectory latent}
The fixed-observation intervention and matched no-VTL comparison reveal
complementary roles for language and VTL. Language changes the requested axial
mode under the same visual state, whereas VTL improves directional correctness
and closed-loop retraction despite not increasing the magnitude of the
instruction-induced response. This suggests that VTL primarily improves
trajectory execution rather than mode selection.

\paragraph{Limitations}
The evaluation uses phantoms with ten trials per method--condition cell and ten
ex-vivo porcine-trachea trials. These limitations motivate
larger repeated closed-loop studies with synchronized video, action, trigger,
and force measurements in the future work.

\FloatBarrier
\section{Conclusion}
\label{sec:conclusion}

We formulated the switch between forward navigation and triggered retraction
as intent aliasing and introduced \methodname{} for bidirectional endoscopic
control. It is a PaliGemma~2-based VLA with a 32-D variational trajectory latent
that conditions a rectified-flow Transformer action expert for receding-horizon
continuous endoscope control. A separate external trigger supervisor switches the instruction outside the learned network.
The architecture-controlled ModeFlag reference shows that a discrete mode interface is viable when
an upstream selector already resolves language. In the paired mode-switch
evaluation, canonical text produces a larger canonical paired response, while the
text-conditioned policy retains 82.8\% IFA across 44 held-out surface forms.
Relative to \rfwithoutlatent{}, \methodname{}
improves navigation-direction accuracy by 11.1 percentage points, reduces
wrong-direction advance during retraction by 83\%, and raises closed-loop
success by 30 percentage points in both the seen colon domain and the held-out
lung and stomach phantoms. The latter serve as out-of-domain luminal transfer
tests beyond the training-domain colon geometry. \methodname{} also completes
all ten ex-vivo porcine-trachea trials following the human-triggered instruction
switch.

\IEEEtriggeratref{40}
\bibliographystyle{IEEEtran}
\bibliography{refs}

\end{document}